\pdfoutput=1

\documentclass[11pt]{article}

\usepackage[final]{acl}

\usepackage{times}
\usepackage{latexsym}

\usepackage[T1]{fontenc}

\usepackage[utf8]{inputenc}

\usepackage{microtype}

\usepackage{inconsolata}

\usepackage{graphicx}

\usepackage{amsmath}
\usepackage{amsthm}
\usepackage{booktabs}
\usepackage{algorithm}
\usepackage{algorithmic}
\usepackage{multirow}
\usepackage{multicol}
\usepackage{amssymb}
\usepackage{makecell}
\usepackage{wrapfig}
\newcommand{\keypoint}[1]{\textbf{#1}}

\title{VLASCD: A Visual Language Action Model for Simultaneous Chatting and Decision Making}

\author{
 \textbf{Zuojin Tang\textsuperscript{1,2}},
 \textbf{Bin Hu\textsuperscript{3}},
 \textbf{Chenyang Zhao\textsuperscript{4}},
 \textbf{De Ma\textsuperscript{1}},
 \textbf{Gang Pan\textsuperscript{1}},
 \textbf{Bin Liu\textsuperscript{2}\thanks{Corresponding Author}}
\\
 \textsuperscript{1}College of Computer Science and Technology, Zhejiang University\\
 \textsuperscript{2}Home Robotics Lab, E-surfing Digital Life Technology Co., Ltd., China Telecom\\
 \textsuperscript{3}Zhejiang Lab,
 \textsuperscript{4}Trinity College Dublin\\
 \texttt{\{zuojintang, made, gpan\}@zju.edu.cn}\\
 \texttt{hubin@zhejianglab.com; zhaoc4@tcd.ie; bins@ieee.org}
}

\begin{document}
\maketitle

\begin{abstract} Recent large pretrained models such as LLMs (e.g., GPT series) and VLAs (e.g., OpenVLA) have achieved notable progress on multimodal tasks, yet they are built upon a multi-input single-output (MISO) paradigm. We show that this paradigm fundamentally limits performance in multi-input multi-output (MIMO) scenarios, where parallel task execution is required. In MISO architectures, tasks compete for a shared output channel, creating mutual exclusion effects that cause unbalanced optimization and degraded performance. To address this gap, we introduce MIMO-VLA (VLASCD), a unified training framework that enables concurrent multi-task outputs, exemplified by simultaneous dialogue generation and decision-making. Inspired by human cognition, MIMO-VLA eliminates interference between tasks and supports efficient parallel processing. Experiments on the CARLA autonomous driving platform demonstrate that MIMO-VLA substantially outperforms state-of-the-art MISO-based LLMs, reinforcement learning models, and VLAs in MIMO settings, establishing a new direction for multimodal and multitask learning. Our code is available at: \textit{https://github.com/Mark-zjtang/MIMO-VLA}
\end{abstract}

\section{Introduction}
Since the emergence of ChatGPT, large language models (LLMs) have become the prototypical large-scale pretrained models. Trained on vast corpora of text and code, they capture rich world knowledge and exhibit strong generalization, including in-context learning and reasoning (e.g., chain-of-thought \cite{wei2022chain}). A growing trend is the extension of LLMs beyond language tasks such as dialogue and text generation toward decision-making in open physical environments.

Currently, three main paradigms exist for building large-scale pretrained models for decision-making in open physical environments. (1) Sequence modeling: methods such as Decision Transformers \cite{chen2021decision} serialize the decision-making process, treating actions as tokens analogous to language; however, they rely heavily on large, high-quality decision-making datasets. (2) Hierarchical modular systems: LLMs perform high-level planning \cite{chen2024driving,carta2023grounding,hu2023enabling,zhou2023large}, decomposing tasks and coordinating specialized modules or tools. (3) End-to-end Vision-Language-Action (VLA) models: approaches like \cite{padalkar2023open,kim2024openvla} bypass modularity altogether, directly mapping multimodal inputs to action outputs.

In recent years, pretrained large models have achieved remarkable progress on multimodal tasks, with LLMs (e.g., ChatGPT) and VLAs (e.g., OpenVLA) as representative examples. These models typically follow a multi-input single-output (MISO) architecture, where multiple inputs produce a single output, and have shown strong performance in text generation and visual understanding.

However, our investigation reveals fundamental limitations of MISO architectures in multi-input multi-output (MIMO) scenarios, such as parallel multi-task execution. Existing MISO LLMs \cite{chen2024driving,liu2023chatgpt} often fail to generate effective actions without compromising dialogue, while MISO VLAs \cite{kim2024openvla,liu2024visual} struggle to produce coherent dialogue at all. This stems from task interference: competing tasks contend for shared output channels, leading to imbalanced optimization and degraded performance. By contrast, human cognition inherently supports non-interfering concurrent task execution (e.g., dialogue and decision-making).

Motivated by this gap, we propose MIMO-VLA (VLASCD), a unified MIMO training architecture with parallel multi-task output capabilities, instantiated as a visual-language-action model for simultaneous dialogue and decision-making. Evaluated on the CARLA 0.9.10 autonomous driving platform \cite{dosovitskiy2017carla}, MIMO-VLA demonstrates that task-adaptive distributed output mapping not only enables efficient multimodal collaboration but also resolves the interference bottleneck of existing MISO models.

The main contributions of this work are:
\begin{itemize}
\item We provide the first evidence that existing MISO models (e.g., LLMs and VLAs) fundamentally fail to handle MIMO tasks.
\item We propose MIMO-VLA, a unified MIMO training architecture with parallel multi-task output capabilities. The design integrates several validated techniques:
(1) a computational module and loss term for generating continuous action values;
(2) an image reconstruction loss to exploit rich visual information during text generation and decision-making;
(3) a label smoothing strategy to preserve dialogue capabilities while enhancing decision accuracy.
\item Extensive experiments demonstrate that MIMO-VLA achieves more accurate real-time action decisions than state-of-the-art (SOTA) baselines, while fully retaining real-time dialogue functionality.
\end{itemize}
\begin{figure*}
    \centering
    \includegraphics[width=0.95\linewidth]{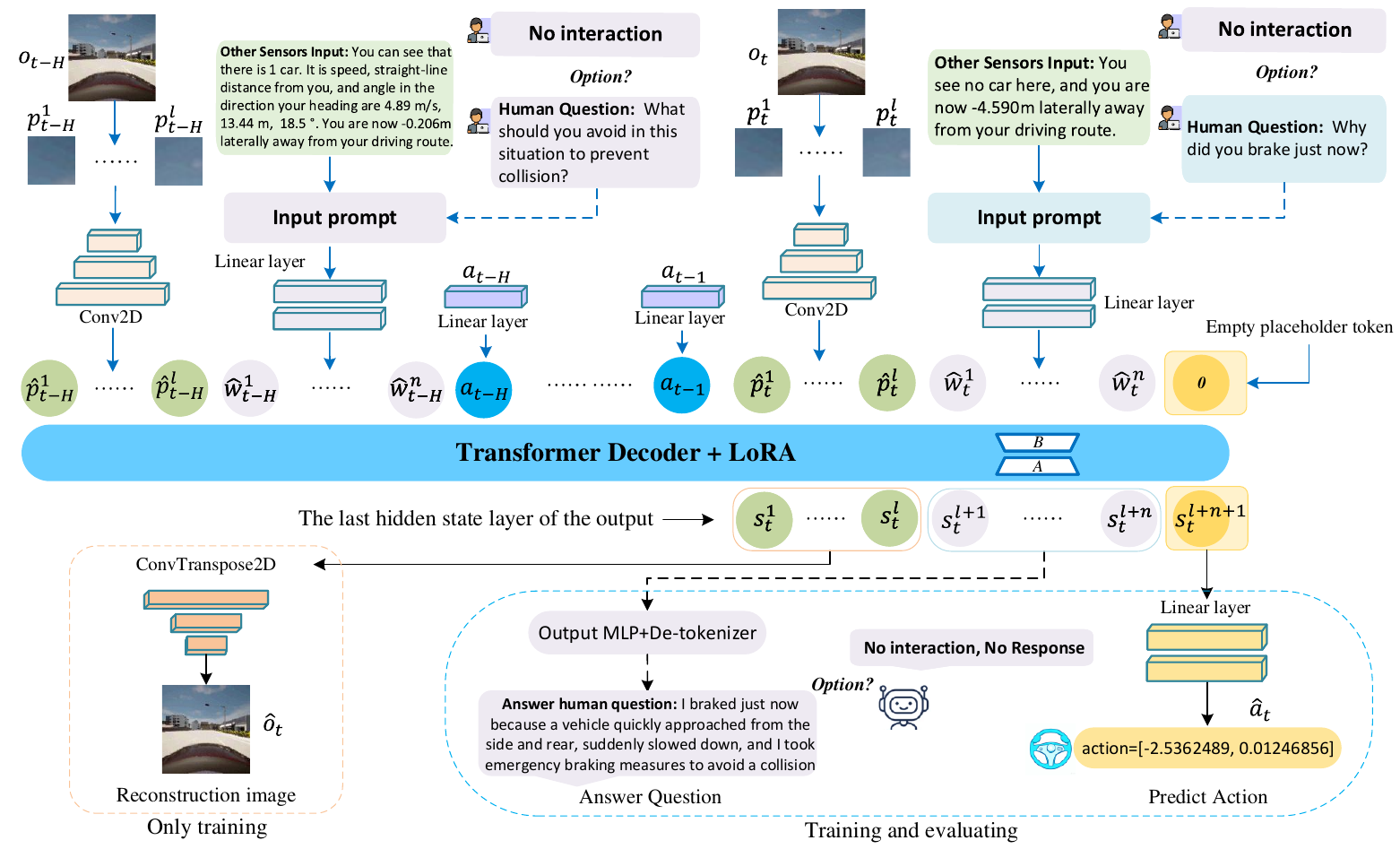}
    \caption{Overview of the MIMO-VLA framework. Expert dataset images, text descriptions, and action values are first mapped into feature representations through linear encoders. These representations are concatenated in a fixed order and fed into a transformer backbone. In the final layer of the LoRA-tuned model, we (i) reconstruct sensor outputs (training only), (ii) generate query responses, and (iii) map continuous actions (during both training and evaluation).}
    \label{fig:overview}
\end{figure*}
\section{Related Work}
\subsection{LLMs for decision-making}
Since \cite{brown2020language}, GPT has established itself as the dominant paradigm for LLMs. Successive models such as GPT-3.5 and GPT-4 \cite{openai2023gpt} have demonstrated strong zero-shot generalization and reasoning. The open-source LLaMA series \cite{touvron2023llama,touvron2023llama2} further accelerated progress, while advances such as chain-of-thought reasoning \cite{wei2022chain} and ReAct \cite{yao2022react} improved reasoning and action generation. More recently, LLMs have been integrated into hierarchical modular decision-making agents, where they typically perform high-level planning rather than directly generating decisions \cite{ahn2022can,fu2023driving,carta2023grounding,chen2024driving,xu2024drivegpt4,sha2023languagempc,hu2023enabling,zhou2023large}.

In contrast, the proposed MIMO-VLA can be viewed as a multimodal GPT variant fine-tuned for downstream application scenarios, distinguished by its ability to simultaneously output action decisions and natural-language dialogue.
\subsection{VLA model for decison-making}
Vision-Language-Action (VLA) models integrate multimodal inputs for embodied decision-making. Unlike conversational LLMs such as ChatGPT, VLAs generate control signals for physical agents (e.g., robots) interacting with the environment. By combining language understanding, visual perception, and action generation \cite{huang2023embodied,li2023vision,zhen20243d,dorka2024matters}, they excel at instruction-following tasks. Compared to deep reinforcement learning (RL) approaches, VLAs demonstrate superior versatility, flexibility, and generality in complex environments \cite{padalkar2023open,brohan2023rt,rfm,team2024gemma,li2023textbooks,bai2023qwen,li2022blip,li2023blip,liu2024visual,tan2019lxmert}.

However, leading VLA models such as RT-X \cite{padalkar2023open} and OpenVLA \cite{kim2024openvla} typically discretize continuous action spaces into fixed intervals. This discretization imposes a critical limitation, preventing them from handling the fine-grained continuous actions required for nuanced operations in complex tasks.
\subsection{LLMs for MIMO settings}
Existing multitasking approaches often rely on task-specific designs, which hinder collaborative optimization and increase computational cost~\cite{geng2022recommendation,liu2023chatgpt,ouyang2022training,driess2023palm}. Recent efforts toward unified frameworks—such as multitask fine-tuning with CGC-LoRA for LLMs~\cite{song2024framework} or generalized policies for multi-task learning~\cite{driess2023palm,alayrac2022flamingo,reed2022generalist,jiang2022vima,ahn2022can}—remain constrained to MISO architectures and thus cannot produce MIMO outputs. The most recent Simlingo framework~\cite{renz2025simlingo} shares certain conceptual similarities, but our work differs in three key aspects:
(1) we identify and analyze multi-task interference in MISO models;
(2) we introduce end-to-end continuous action mapping;
(3) we design a dynamic composite loss integrating language, action, and image objectives to enable more efficient multimodal learning.
\section{Methodology}
In this section, we present how to build MIMO-VLA in detail, including the model architecture and the training procedure, with a focus on the loss designs in the last output hiddens layer. An overview of MIMO-VLA is illustrated in Figure~\ref{fig:overview}. To begin with, we present the problem setting of our concern.
\subsection{Problem Setting}
We consider a multimodal setting similar to \cite{xiao2020multimodal}, where at each time step $t$, the agent executes an action $a_t$ and the environment returns visual and textual observations, denoted $\{o_t, \hat{w}_t\}$. Our goal is to learn a generative model $\pi(\hat{a}_t, \hat{w*}_t|o_{t-H}, \hat{w}_{t-H}, a_{t-H}, ..., o_t, \hat{w}_t)$ that produces both high-quality action decisions $\hat{a}_t$ and textual responses $\hat{w*}_t$ conditioned on a sequence of past trajectories. Here, $H$ denotes the context length.
\subsection{Model Architecture}
Our model supports three input modalities: text, image, and numeric vectors. We adopt LLaMA-7B \cite{touvron2023llama2} as the backbone and encode textual inputs using its pretrained embedding layers. Visual inputs are encoded following standard practices in visual-language models (VLMs) \cite{liu2024visual} and VLAs \cite{kim2024openvla}: each image $o_t$ is first divided into $L$ patches $p_l, l=1,\ldots,L$, which are then mapped into the vector space via a trainable 2D convolutional network. For numeric action inputs, a multi-layer perceptron (MLP) encodes action values into the same vector space. Finally, embeddings from all three modalities are concatenated to form the trajectory embedding sequence $\tau$ at time $t$, defined as follows:
\begin{multline}
\textstyle
\tau_t = \{(\hat{p}^1_{t-H},\ldots,\hat{p}^L_{t-H}), (\hat{w}^1_{t-H},\ldots,\hat{w}^n_{t-H}), \\ a_{t-H}, \ldots, (\hat{p}^1_t,\ldots,\hat{p}^l_t), (\hat{w}^1_t,\ldots,\hat{w}^n_t)\}
\end{multline}
where $\hat{p}_{t}^i$ and $\hat{w}_{t}^j$ denote the embeddings of the $i$-th visual patch and the $j$-th textual token at time step $t$, respectively.

During inference, the transformer backbone of MIMO-VLA generates hidden embeddings $s_t^{l+1},\cdots,s_t^{l+n+1}$ (Figure~\ref{fig:overview}), which are subsequently decoded into outputs for different modalities. MIMO-VLA supports two output modalities: textual responses for dialogue and numeric vectors for action-level decision-making. For dialogue, we leverage the pretrained output MLP layers and tokenizer of LLaMA-7B to generate text. For action decisions, the model generates an additional embedding vector following the ``$<\text{EOS}>$''placeholder token. Unlike prior approaches such as OpenVLA \cite{kim2024openvla} and RT-X \cite{brohan2023rt}, which treat action prediction as a token generation task by discretizing the action space into bins, we employ an action head comprising multiple MLP modules that directly map embeddings to continuous action values. We empirically find that this design achieves superior performance compared to discretization-based methods.
\subsection{Training Procedure}
\label{sec: training}
We fine-tune the transformer backbone using LoRA \cite{hu2021lora} and jointly train the image, text, and action encoding modules, as well as the decoding modules, on an offline dataset $D_{expert}$ containing demonstrated driving trajectories paired with question-answer (QA) annotations. The model learns to predict control actions and respond to driving-related queries, e.g., "\textit{Summarize the current driving scenario}." An auxiliary image reconstruction task is incorporated, where a transposed convolution layer reconstructs input image patches from the output embeddings $s_t^1,\cdots,s_t^l$ to enhance feature learning. The overall training objective comprises three loss terms—text generation, action prediction, and image reconstruction—where $\phi$ denotes decoder parameters and $\theta$ represents all other trainable parameters.

\keypoint{Text Generation.} In our experiments, we observed that simply replacing specific numerical values in the translation template \cite{chen2024driving} produces minimal representational differences due to the sequential nature of the data. Consequently, using conventional cross-entropy loss for text generation can easily lead to model overfitting (see Appendix~\ref{sec:overfitting} for details). To address this, we adopt label smoothing \cite{szegedy2016rethinking} to regularize training. Specifically, the hard label for token $w_i$ is softened by distributing a small portion of the probability mass to incorrect classes:
\begin{equation}
\textstyle
q_i^k = \begin{cases}
1 - \epsilon & \text{if } k = y_i, \\
\frac{\epsilon}{K-1} & \text{otherwise},
\end{cases}
\end{equation}
where $\epsilon$ is the smoothing factor and $K$ is the number of total classes, i.e., the vocabulary size. Accordingly, the final loss term used for text generation is:
\begin{equation}
\textstyle
\mathcal{L}_{\text{language}}(\theta) = \frac{1}{N}\sum_i \sum_k q_i^k \log{p(k|\tau^{:i-1}, \theta)},
\end{equation}
where $\tau^{:i-1}$ denotes the input token sequence preceding position $i$, used to predict token~$i$, and $N$ represents the maximum padding length for input text normalization.

\keypoint{Action Prediction.} To directly predict continuous action values instead of discretized action bins, we train the model using a mean squared error (MSE) loss between the ground-truth action $a_t$ and the predicted action $\hat{a}_t$, defined as follows:
\begin{equation}
\textstyle
 \mathcal{L}_{\text{action}}(\theta) = \frac{1}{T} \sum_{t=1}^T \frac{1}{D}\sum_{d=1}^D [(a_t^d - \pi_{\theta}(\hat{a}_t^d|\tau_t))^2]
\end{equation}
where $T$ denotes the sequence length per training sample. Where $D$ denotes the dimensionality of the action space. In our experiments, $D=2$, corresponding to vehicle acceleration and steering.
\begin{figure*}
    \centering
    \includegraphics[width=0.95\linewidth]{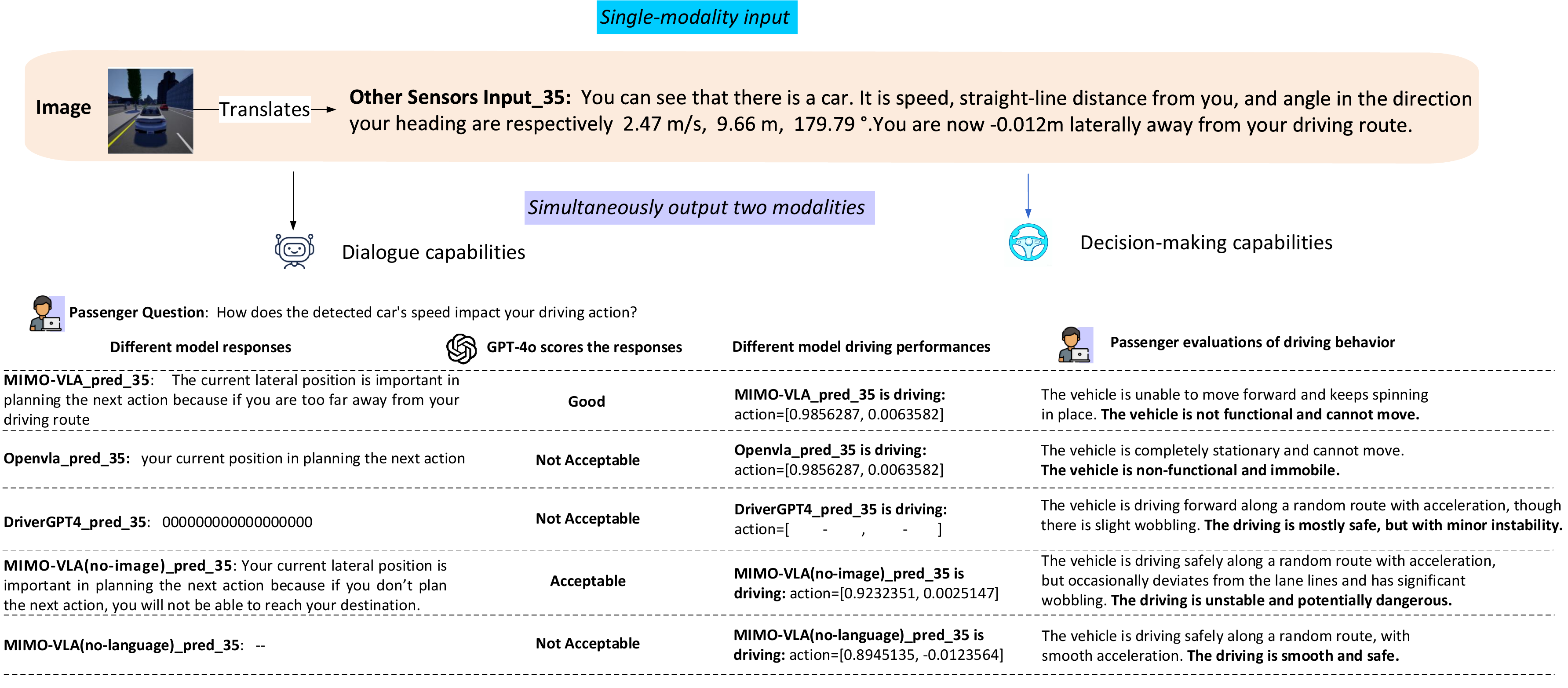}
     \caption{Random examples of different MIMO models engaging in fluent human–machine dialogue while simultaneously making real-time driving decisions.}
    \label{fig:enter-label}
\end{figure*}

\keypoint{Image Reconstruction.} To better exploit the rich environmental information in visual inputs while mitigating information loss under limited data, we introduce an auxiliary image reconstruction task. This provides additional supervision for the visual modality by using a 2D transposed convolution layer $f_\phi$ to reconstruct image patches from their embeddings. The reconstruction loss is defined as the pixel-wise Euclidean distance between the original and reconstructed patches:
\begin{equation}
\textstyle
    \mathcal{L}_{\text{image}}(\theta, \phi) = \frac{1}{L}\sum_{l=1}^L\text{mse}(o_t, f_\phi(\pi_{\theta}(g_\theta(\tau_t^{:p_t^l}))))
\end{equation}
where $o_t$ denotes the input image, $\tau_t^{:p_t^l}$ represents the input sequence up to patch $p_t^l$, and $g_\theta$ is a trainable 2D convolutional network that maps image patches $p_t^1,\cdots,p_t^l$ directly into the language embedding space $\hat{p}_t^1,\cdots,\hat{p}_t^l$. 

\keypoint{Loss Function for Model Training}
In summary, the overall training loss for our model is defined as:
\begin{equation}
\textstyle
\mathcal{L} = \alpha_1\mathcal{L}_{\text{language}}+\alpha_2\mathcal{L}_{\text{action}}+\lambda\mathcal{L}_{\text{image}}
\end{equation}
where $\alpha_1, \alpha_2, \lambda$ are the weighting hyperparameters for the three loss components. Importantly, we explicitly implement a gradient-space isolation mechanism: $\mathcal{L}_{\text{language}}$ affects only text token positions, $\mathcal{L}_{\text{action}}$ affects only action token positions, and $\mathcal{L}_{\text{image}}$ affects only image patch positions. Ablation studies (Table 4) systematically validate the independent contribution of each component, demonstrating that this design effectively encourages the formation of modality-specific representations at distinct positional encodings within the model.
\section{Experiments}
In this section, we validate MIMO-VLA on the CARLA autonomous driving simulation platform, demonstrating its ability to make fine-grained action decisions while maintaining dialogue capabilities. Our experiments focus on: (1) the impact of individual loss components on model performance, and (2) the influence of textual data quality on driving decisions.
\subsection{Experimental setting}
Our experiments were conducted in gym-carla \cite{gym-carla}, an OpenAI Gym-compatible environment built on CARLA 0.9.10. For LoRA fine-tuning, we selectively updated only the $Q$ and $V$ projection modules, comprising 0.06\% of LLaMA-7B’s parameters. Additional implementation details—including MIMO-VLA hyperparameters, linear mapping layers, and gym-carla configurations—are provided in Appendix~\ref{sec:hyperparameter}.
\subsection{Comparison methods}
Behavior Cloning (BC) implemented in gym-carla was used as a baseline. Comparative methods include the RL approaches Dreamer \cite{hafner2019dream} and Forbes \cite{chen2022flow}, Decision Transformer (DT) \cite{chen2021decision}, and VLA models OpenVLA \cite{kim2024openvla} and DriverGPT4 \cite{xu2024drivegpt4}.
\subsection{Training datasets}
\begin{figure}
    \centering
    \includegraphics[width=1.0\linewidth]{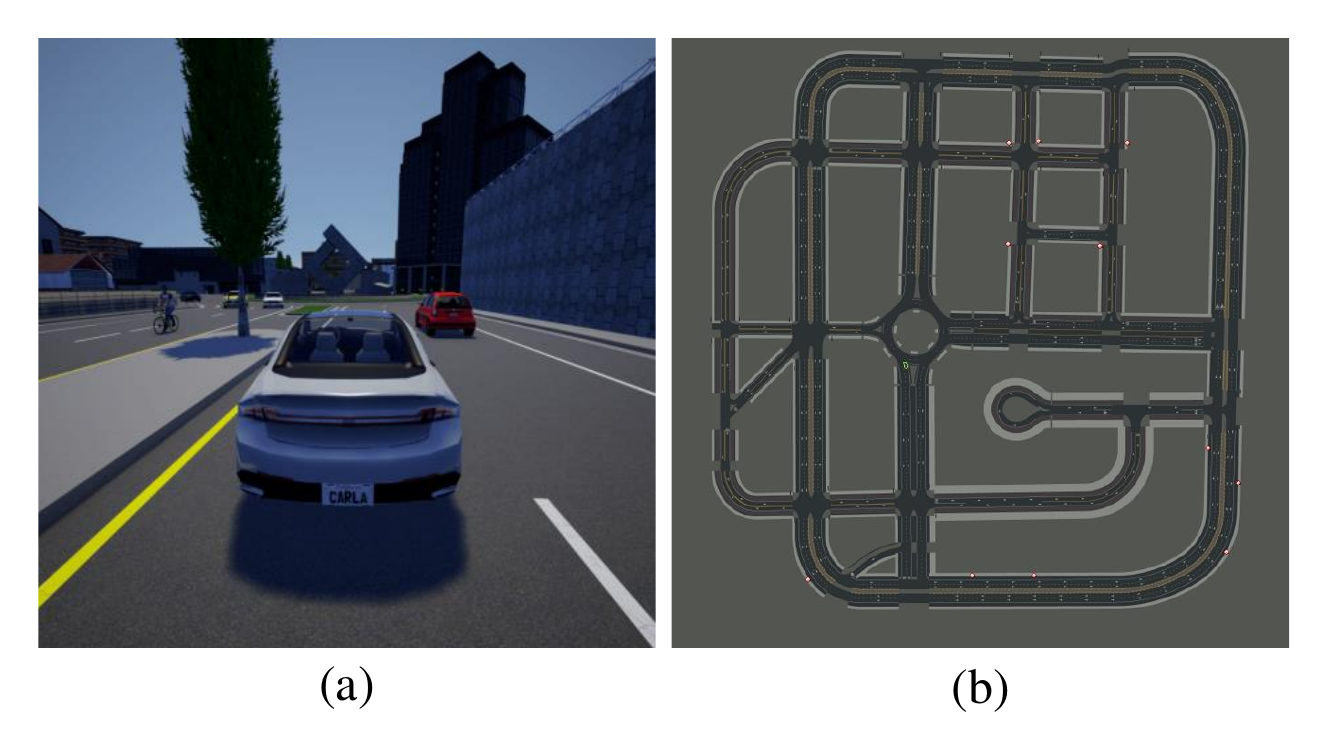}
    \caption{(a) a sample view of the simulation environment; (b) a bird-eye view of our task scenario.}
    \label{fig:carla}
\end{figure}

The training dataset $D_{expert}$ was collected using the EGADS framework \cite{tang2024efficient}, which designs RL- and IL-based agents with safety constraints and demonstrates strong performance in CARLA. We use this agent as our expert, driving vehicles in town03 of CARLA to collect the dataset. $D_{expert}$ totals 5.69GB and contains 13,761 frames. For each frame, one question from a set of 50 was randomly selected based on the textual description of the current observation, along with its corresponding answer. The 50-question set is detailed in Appendix~\ref{sec: problems set}, and further descriptions of $D_{expert}$ and the map are provided in Appendix~\ref{sec:traning_datasets} and Appendix~\ref{sec:CARLA_maps}, respectively. Figure~\ref{fig:carla}(b) illustrates the layout of town03 used for training.

During data collection and online evaluation, vehicles randomly select directions at intersections, follow randomly generated routes, slow down for preceding vehicles, and stop at red lights. Actions are defined as \text{action} = [\text{accel}, \text{steering}], with acceleration ranging from [-3, 3] ([0, 3] for acceleration, [-3, 0] for deceleration) and steering range in [-0.2, 0.2]. Additional implementation details are provided in Appendix~\ref{sec:hyperparameter}.

Following \citet{chen2024driving}, we design a template-based parser that converts sensor data (e.g., position and distance information, excluding vision and LiDAR) into natural language descriptions, shown as “other sensors input” in Figures 1 and 2. This input excludes any action-related information from MIMO-VLA, such as speed and heading, allowing us to evaluate whether the model can leverage textual information to improve action decision quality. Details on the templates are provided in Appendix~\ref{sec:template}.
\subsection{Performance metrics}

\keypoint{Performance Metrics for Evaluating Chatting Ability}
The study by \citet{wang2023chatgpt} demonstrates that ChatGPT achieves high alignment with human judgments. Building on this finding, we employed GPT-4o \cite{openai2023gpt} to systematically compare the answer quality of MIMO-VLA with baseline models. Our evaluation procedure consisted of three steps: (1) selecting 50 random driving environment–question pairs; (2) generating responses from the baseline models for each pair; and (3) scoring the responses on a 0–10 scale using GPT-4o, with the following categories: Not Acceptable ($<3$), Acceptable ($3 \leq \text{score} < 6$), and Good ($\geq 6$). The full evaluation prompt is provided in Appendix~\ref{sec:template}.

To further examine the contributions of the language and image components to dialogue capabilities, we evaluated two simplified variants of MIMO-VLA: MIMO-VLA (no-language) and MIMO-VLA (no-image). These variants were trained by removing the loss terms corresponding to text generation and image reconstruction, respectively, resulting in $\mathcal{L}{\text{action}} + \mathcal{L}{\text{image}}$ and $\mathcal{L}{\text{action}} + \mathcal{L}{\text{language}}$.

\keypoint{Performance metrics for evaluating the decision-making ability}
We deployed our trained model on a vehicle for autonomous urban navigation and evaluated its performance using established metrics: Collision Rate (CR), Off-road Rate (OR), Episode Completion Rate (ER), Average Safe Driving Distance (ASD), Average Reward (AR), and Driving Score (DS). DS, a composite metric reflecting overall performance, is defined as DS = ER $\times$ AR, following the methodology of the CARLA Leaderboard. For AR, we adopted the reward function from \citet{chen2019model}, which accounts for driving dynamics such as yaw, collisions, speeding, and lateral velocity. Model selection prioritized checkpoints that optimized both DS and AR. The remaining metrics (ER, OR, ASD) were implemented in accordance with \citet{gao2024enhance}. Additional details on reward computation and metric calculations are provided in Appendices~\ref{sec:app-reward} and~\ref{sec:metrics}.
\begin{figure}
    \centering
    \includegraphics[width=1.0\linewidth]{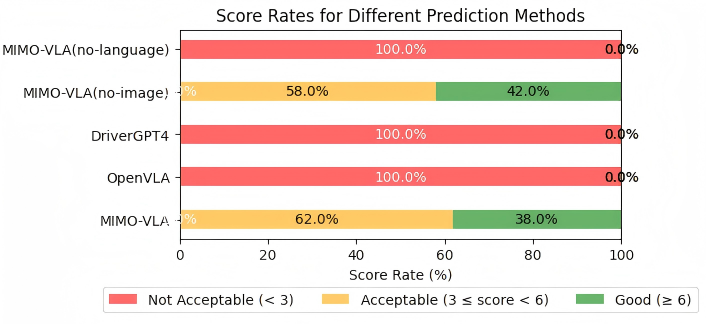}
    \caption{GPT-4o scores the answers from five methods for randomly generated inputs and question}
    \label{fig:gpt-4o}
\end{figure}
\begin{table*}[t!]
\centering
\caption{Evaluation results for different methods in town03 (random), \textit{H=1}}
\resizebox{135mm}{!}{
\begin{tabular}{cccc|cccc}
\toprule
Method & Input &DS $\uparrow$ & AR ($f$) $\uparrow$ & ASD(m) $\uparrow$ & ER(\%) $\uparrow$ & OR(\%) $\downarrow$ & CR(\%) $\downarrow$    \\
\midrule
BC & image & $20.21\pm 7.46$ & $175.34\pm72.86$ & $54.21\pm 6.41$ & $9.08\pm 0.56$ & $54.86\pm 20.04$ & $60.00\pm 11.23$ \\
\midrule
DriverGPT4&image, text &- & - & - & - & - &- \\
Openvla&image, text & $-13.02\pm 4.02$ & $-199.16\pm38.73$ & $24.34\pm 5.02 $& $5.25\pm 0.39 $& $24.36\pm 4.17$ & $95.00\pm 0.00$ \\
MIMO-VLA&image, text&$92.78\pm23.75$ &$466.80\pm91.66 $&$71.77\pm9.40$ &$ 16.35\pm 1.56$ &$15.33\pm4.36$  &$55.00\pm11.41$ \\

\bottomrule
\end{tabular}}
\label{tab:table1}
\end{table*}

\begin{table*}[t!]
\centering
\caption{Evaluation results for different methods in town03 (random), \textit{H=4}}
\resizebox{135mm}{!}{
\begin{tabular}{cccc|cccc}
\toprule
Method & Input& DS $\uparrow$ & AR ($f$) $\uparrow$  & ASD(m) $\uparrow$ & ER(\%) $\uparrow$ & OR(\%) $\downarrow$ & CR(\%) $\downarrow$  \\
\midrule
BC& image& $36.39\pm 13.37$ & $314.66\pm86.02$ & $64.08\pm10.48$ & $9.04\pm 0.62$ & $37.56\pm16.44$& $45.00\pm 11.41$\\
Dreamer& image &$ -0.03\pm 0.01$ & $-14.96\pm0.09$ & $0.02\pm0.01 $&$0.22\pm 0.01$ &$ 0.00\pm0.00$& $0.00\pm 0.00$ \\
Forbes& image & $0.98\pm 1.43$ &$ 21.63\pm21.72$ & $22.84\pm1.00 $&$6.30\pm 0.31$ & $18.78\pm1.03$& $56.67\pm 9.20$ \\
DT& image&$ 7.68\pm3.24$ & $51.97\pm29.33 $&$23.74\pm2.47$  & $9.92\pm0.71 $&$10.31\pm2.32$ &$65.00\pm10.94$\\
\midrule
DriverGPT4&image, text & - & - & - & - & - & - \\
Openvla& image, text &$ -7.84\pm 0.67$ & $-160.37\pm7.85$ & $18.03\pm1.92$ & $4.76\pm 0.19 $& $20.77\pm3.36$& $100.00\pm 0.00 $\\
MIMO-VLA & image, text &$105.25\pm14.03$  & $349.52\pm49.75$ & $59.76\pm\ 5.04 $ &$25.02\pm2.57 $&$19.93\pm2.11$ &$30.00\pm10.51$\\
\bottomrule
\end{tabular}}
\label{tab:table2}
\end{table*}

\begin{table*}[t!]
\centering
\caption{Evaluation the generalization for different methods in town04 (random), \textit{H=4}}
\resizebox{135mm}{!}{
\begin{tabular}{cccc|cccc}
\toprule
Method & Input& DS $\uparrow$ & AR ($f$) $\uparrow$  & ASD(m) $\uparrow$ & ER(\%) $\uparrow$ & OR(\%) $\downarrow$ & CR(\%) $\downarrow$  \\
\midrule
BC& image& $39.22\pm11.64$  & $358.79\pm79.59$ &$63.08\pm9.37 $&$8.69\pm 0.56 $&$ 5.64\pm1.26$&$60.00\pm 11.23$\\
Dreamer& image &$ -0.03\pm0.01 $& $-15.03\pm0.07$ & $0.02\pm0.01 $&$0.01\pm 0.21$ & $0.01\pm0.00$&$ 0.00\pm 0.00$ \\
Forbes& image &$-2.63\pm 2.75$ & $-17.37\pm22.98$ & $19.79\pm1.20 $&$6.24\pm 0.69$ & $15.80\pm2.74$& $66.70\pm 8.75$ \\
DT& image&$10.66\pm 3.26 $ &$85.58\pm 27.04 $ &$24.94\pm2.92$  &$10.55\pm 0.58$ &$11.38\pm 2.15$ &$55.00\pm 11.41$\\
\midrule
DriverGPT4&image, text & - & - &- & - & - &-\\
Openvla& image, text & $-6.74\pm 0.88$ & $-153.35\pm10.26$ & $13.62\pm1.86 $& $4.26\pm 0.17$ &$ 15.70\pm2.71$& $100.00\pm 0.00$ \\
MIMO-VLA& image, text & $94.26\pm15.26 $& $384.52\pm51.72$ &$56.93\pm4.03 $ &$21.49\pm1.86$ &$12.75\pm2.28$ &$45.00\pm11.41$\\

\bottomrule
\end{tabular}}
\label{tab:table3}
\end{table*}

As shown in Figures 2 and 4, MIMO-VLA significantly outperforms the other models in terms of chatting ability. In contrast, OpenVLA performs poorly in question-answering, as it focuses solely on optimizing the action loss. DriverGPT4 also struggles: since both tasks share the same decoder, the model tends to interpret all inputs as intended for action prediction, making it difficult to generate complete text. Although DriverGPT4 has two independent loss terms, it fails to effectively balance them.

Ablation results further highlight the role of different components. MIMO-VLA (no-language) exhibits a substantial drop in conversational performance compared to the full MIMO-VLA, whereas MIMO-VLA (no-image) performs comparably to MIMO-VLA. This indicates that the language loss component is crucial for enhancing chatting capabilities, while the image loss contributes less directly to dialogue performance.
\subsection{Results on Decision-Making Ability Evaluation}
In Tables 1–3, the "-" entries indicate system failure cases where no complete action value is generated within the 50-second threshold, representing the maximum allowable stationary wait time before triggering system intervention. Here, $H$ denotes the length of the context.

As shown in Table 1, MIMO-VLA significantly outperforms BC and OpenVLA in DS, AR, and ASD at a single time step, while DriverGPT4 fails to produce precise action values. Across multiple time steps (Table 2), MIMO-VLA continues to show substantial improvements over other methods, indicating sustained performance over longer durations.

To evaluate generalization, models were trained on the town03 dataset and tested online in town04. Table 3 demonstrates that MIMO-VLA achieves a markedly higher DS than the other methods, highlighting its strong generalization capability. By contrast, DriverGPT4 struggles to generate precise action values for real-time control commands, reflecting the challenges of producing accurate outputs using a detokenizer. OpenVLA can generate precise values in experiments but often produces identical action commands, causing the vehicle to spin or wander in place, incurring significant penalties.

Additionally, Tables 2 and 3 show that MIMO-VLA significantly outperforms DT, Dreamer, and Forbes in both decision-making performance and generalization. Finally, Figure 2 illustrates that MIMO-VLA can seamlessly engage in conversation with a human while simultaneously making real-time driving decisions.
\begin{table*}[t!]
\centering
\caption{Ablation Study on MIMO-VLA Loss Functions in Town03 (Random), with \textit{H = 4}}
\resizebox{135mm}{!}{
\begin{tabular}{cccc|cccc}
\toprule
Loss function & Input& DS $\uparrow$ & AR ($f$) $\uparrow$  & ASD(m) $\uparrow$ & ER(\%) $\uparrow$ & OR(\%) $\downarrow$ & CR(\%) $\downarrow$  \\
\midrule
$\mathcal{L}_{\text{image}}$+$\mathcal{L}_{\text{language}}$+$\mathcal{L}_{\text{action-bins}}$& image, text &$11.57\pm0.00$ & $142.83\pm0.01$ & $22.71\pm\ 0.01$  &$8.10\pm0.05 $&$30.87\pm0.10 $&$100.00\pm0.00$\\
$\mathcal{L}_{\text{image}}$+$\mathcal{L}_{\text{action}}$&$ image, text$ &$45.08\pm10.88$  & $234.36\pm52.21$ & $39.64\pm4.03 $&$14.13\pm1.71$ &$16.68\pm3.15 $&$30.00\pm10.51$\\
$\mathcal{L}_{\text{language}}$+$\mathcal{L}_{\text{action}}$&image, text& $74.85\pm10.97$& $331.78\pm49.88$ &$50.63\pm4.73$& $18.62\pm1.95$&$ 15.96\pm 2.45$ & $25.00\pm9.93$\\
$\mathcal{L}_{\text{image}}$+$\mathcal{L}_{\text{language}}$+$\mathcal{L}_{\text{action}}(our)$& image, text &$105.25\pm14.03 $ & $349.52\pm49.75$ & $59.76\pm\ 5.04 $ &$25.02\pm2.57$ &$19.93\pm2.11$ &$30.00\pm10.51$\\
\bottomrule
\end{tabular}}
\label{tab:table4}
\end{table*}

\begin{table*}[t!]
\centering
\caption{Impact of Sensor Input and QA Content Noise on MIMO-VLA’s Decision-Making Performance in Town03 (Random)}
\resizebox{135mm}{!}{
\begin{tabular}{cccc|cccc}
\toprule
Input& Sensor input / QA (noise ratio) &DS $\uparrow$ & AR ($f$) $\uparrow$  & ASD(m) $\uparrow$ & ER(\%) $\uparrow$ & OR(\%) $\downarrow$ & CR(\%) $\downarrow$   \\
\midrule
image, text & $0\% / 100\%$& $74.32\pm24.44$ & $288.54\pm74.62$ & $62.42\pm7.71$ & $25.76\pm 1.54 $& $11.05\pm 1.62$ & $50.0\pm 0.51$\\
image, text&$ 0\% / 0\% $&$93.89\pm29.73$ &$336.11\pm86.72 $&$45.42\pm9.53$ &$ 16.68\pm 2.50$ &$19.05\pm4.96$ &$5.00\pm5.00$ \\
image, text & $100\% / 0\%$& $-0.01\pm 1.12$ &$ -5.10\pm0.00 $& $0.00\pm 0.00$ &$ 0.30\pm 0.00$ &$ 0.00\pm 0.00$ &$ 0.00\pm 0.00$\\

\bottomrule
\end{tabular}}
\label{tab:example}
\end{table*}

\subsection{Ablation Studies on Loss Function Design}
As shown in Equation (6), our loss function consists of three components: action loss $\mathcal{L}{\text{action}}$, language loss $\mathcal{L}{\text{language}}$, and image loss $\mathcal{L}_{\text{image}}$. We conducted ablation studies to investigate the contribution of each loss to MIMO-VLA’s performance. The results are presented in Table~\ref{tab:table4}.

For comparison, we include the action-bins loss $\mathcal{L}{\text{action-bins}}$ used by OpenVLA and RT2, which handle continuous-valued actions via value discretization. We also evaluate two simplified variants of MIMO-VLA: MIMO-VLA (no-language) and MIMO-VLA (no-image). These were trained using $\mathcal{L}{\text{action}} + \mathcal{L}{\text{image}}$ and $\mathcal{L}{\text{action}} + \mathcal{L}_{\text{language}}$, respectively, to isolate the effects of the language and image loss components.

\textbf{On the effect of $\mathcal{L}_{\text{action}}$} As shown in Table 4, comparing the performance of $\mathcal{L}{\text{image}} + \mathcal{L}{\text{language}}+\mathcal{L}{\text{action-bins}}$ with $\mathcal{L}{\text{image}}+\mathcal{L}{\text{language}}+\mathcal{L}{\text{action}}$ clearly demonstrates the advantage of our action loss $\mathcal{L}{\text{action}}$. This explains why MIMO-VLA outperforms VLA models using action losses similar to $\mathcal{L}{\text{action-bins}}$, as observed in Tables 1–3.

Our experiments indicate that action discretization and tokenization, as used in current VLA models, often yield low training loss but poor inference performance. This occurs because adjacent action intervals are represented by consecutive token IDs that are close in token space, causing the model to repeatedly output the same token during inference, even though the corresponding action values may differ substantially. For example, in OpenVLA, discretizing throttle control (0–100\%) into 256 bins produces intervals of ~0.39\%, allowing only fixed values such as 0\%, 0.39\%, 0.78\%, etc. Token IDs like 31830 and 31831 may be treated by the model as equivalent, preventing fine-grained distinctions.

In contrast, our approach uses an MLP to map actions to continuous control values, enabling direct outputs with arbitrary precision.

\textbf{On the effect of $\mathcal{L}_{\text{language}}$}\quad As shown in Table 4, comparing $\mathcal{L}{\text{image}} + \mathcal{L}{\text{action}}$ (MIMO-VLA (no-language)) with $\mathcal{L}{\text{image}} + \mathcal{L}{\text{language}} + \mathcal{L}{\text{action}}$ (MIMO-VLA) reveals that including $\mathcal{L}{\text{language}}$ in the loss function significantly improves decision-making quality. Figures 2 and 4 further show that MIMO-VLA (no-language) exhibits markedly weaker dialogue capabilities than MIMO-VLA, whereas MIMO-VLA (no-image) performs comparably to MIMO-VLA. These results highlight that $\mathcal{L}{\text{language}}$ is crucial for maintaining strong dialogue performance. In summary, incorporating $\mathcal{L}{\text{language}}$ benefits both dialogue and decision-making capabilities.

\textbf{On the effect of $\mathcal{L}_{\text{image}}$}\quad As shown in Table 4, adding $\mathcal{L}{\text{image}}$ to the loss function (corresponding to $\mathcal{L}{\text{image}} + \mathcal{L}{\text{language}} + \mathcal{L}{\text{action}}$) improves all decision-making performance metrics compared to $\mathcal{L}{\text{language}} + \mathcal{L}{\text{action}}$. This demonstrates that $\mathcal{L}_{\text{image}}$ provides significant benefits for enhancing decision-making. We attribute this improvement to the fact that high-quality image reconstruction allows the model to better extract and leverage rich scene information from the image modality, thereby supporting more informed decision-making.
\subsection{Resolving Conflicts in Simultaneous Multi-Task Output}
In DriveGPT4, the text generation and action generation tasks lack independently designed loss functions, leading to conflicts between them. This is especially problematic in complex environments, where the model struggles to generate both effective text and precise action instructions simultaneously. As shown in Figure 2, DriveGPT4 cannot reliably produce fine-grained action commands at every moment, which compromises both decision accuracy and dialogue quality. In contrast, MIMO-VLA employs separate objective functions for text and action generation, enabling efficient parallel processing of each task and effectively preventing conflicts between them.
\subsection{Impact of Textual Data Quality on Decision-Making Capability}
\label{sec:noise input}
Consider a driver conversing with a passenger while operating a car. If the passenger’s remarks are irrelevant to the driving context, they may interfere with the driver’s decisions, although humans naturally exhibit some tolerance to such noise. To investigate whether our model demonstrates similar human-like decision-making behavior, we designed a set of experiments.

As shown in Table 5, introducing noise unrelated to driving into the sensor inputs causes a rapid decline in the model’s decision-making performance. In contrast, adding noise solely to the QA content while keeping sensor inputs clean results in a less pronounced performance drop. This suggests that the model remains robust as long as the sensor inputs are relevant to the driving scenario, even when the QA content contains noise. These findings indicate that our model’s decision-making behavior closely mirrors that of human drivers.
\section{Conclusion}
In this paper, we investigated how to develop a multimodal pre-training framework to address the inherent task conflicts in MISO architectures under MIMO scenarios, where multiple tasks share output channels. Such conflicts often lead to imbalanced optimization and significant performance degradation in individual tasks. To tackle this, we proposed a unified MIMO training architecture with parallel multi-task output capabilities—MIMO-VLA. Experiments demonstrate that MIMO-VLA outperforms SOTA VLA models, RL approaches, and decision transformers in decision-making while maintaining fluent dialogue, thanks to our continuous-action handling, loss function design, and label smoothing techniques.

The shift from modular systems composed of discrete subcomponents to unified end-to-end models represents a major trend in AI research. In the context of MIMO scenarios, we believe this work constitutes a meaningful first step toward developing a unified generative model capable of simultaneously handling both dialogue and action generation in an end-to-end manner.
\section*{Limitations}
This study has several limitations that guide future research. First, although MIMO-VLA is a general-purpose unified MIMO framework, our validation is limited to autonomous driving, leaving its generalizability to other domains, such as robotics or human-computer interaction, untested. Second, joint optimization of dialogue and action generation still requires improvement, particularly in multi-task coordination and scalability. Our dialogue evaluation also lacks scenario categorization and linguistic diversity, which future studies could address to enable more detailed assessment of conversational capabilities. Finally, long text prompts or large image inputs can create synchronization delays, underscoring the need for more efficient token processing and resource allocation to support real-time performance.
\bibliography{custom}
\clearpage
\appendix

\section{Appendix}
\subsection{Hyperparameter settings}
\label{sec:hyperparameter}
In this section, we present the model parameters of MIMO-VLA, the parameters of the custom linear layers, and the settings for Gym-CARLA and evaluation, as summarized in Tables 6, 7, and 8. The models were trained using Python 3.8, Transformers 4.30.0, and an NVIDIA Tesla V100 GPU. Training times ranged from 5 to 13 hours, depending on the input modality and trajectory length. We also conducted experiments on the three loss function hyperparameters, as detailed in Appendix~\ref{sec: loss weights}, and in our experiments we set $\alpha_1 = 0.1$, $\alpha_2 = 10$, and $\lambda = 0.5$.
\begin{table*}[h!]
    \centering
    \caption{Hyperparameters}
    \begin{tabular}{@{}ll@{}}
        \toprule
        \textbf{Parameter}                & \textbf{Value}    \\ \midrule
        batch\_size                       & 64                \\
        micro\_batch\_size                & 8                  \\
        num\_epochs                       & 3                  \\
        learning\_rate                    & 3e-4               \\
        cutoff\_len                       & 424                \\
        val\_set\_size                   & 0.1                \\
        save\_step                       & 25                 \\
        lora\_r                           & 8                  \\
        lora\_alpha                       & 16                 \\
        lora\_dropout                     & 0.05               \\
        lora\_target\_modules             & \{q\_proj, k\_proj\} \\
        Other Sensors Input\_types                      & \{obs, text\}     \\
        lambda\_action                    & 10                 \\
        lambda\_smooth                    & 0.1                \\
        lambda\_img                       & 0.5                \\
        horizon                           & 1                  \\
        regular\_action\_loss            & False              \\
        img\_patch\_size                 & 16                 \\ \bottomrule
    \end{tabular}
\end{table*}
\begin{table*}[h!]
    \centering
    \caption{Model Parameters and Layers}
    \begin{tabular}{@{}ll@{}}
        \toprule
        \textbf{Parameter/Layer}                      & \textbf{Details}                      \\ \midrule
        num\_patches                                   & 64                     \\
        tokenizer\_vocab\_size                        & 32000                                 \\
        split\_obs\_proj                              & Conv2d(3, 4096, kernel\_size=16, stride=16) \\
        inverse\_split\_obs\_proj                     & ConvTranspose2d(4096, 3, kernel\_size=16, stride=16) \\
        split\_obs\_position\_embedding               & Parameter(torch.randn(1, 64, 4096)) \\
        text\_embedding                                & nn.Embedding(32000, 4096) \\
        custom\_lm\_head                              & Linear(4096, 32000, bias=False) \\
        actor\_linear1                                 & Linear(4096, 2048) \\
        actor\_linear2                                 & Linear(2048, 1024)                    \\
        actor\_linear3                                 & Linear(1024, 512)                     \\
        actor\_linear4                                 & Linear(512, 256)                      \\
        actor\_linear5                                 & Linear(256, 128)                      \\
        actor\_linear6                                 & Linear(128, 64)                      \\
        actor\_linear7                                 & Linear(64, 2)                      \\
        reconstruction\_layer                                 & Linear(4096, micro\_batch\_size*3*128*128)                      \\
        action\_linear                                 & Linear(2, 4096)                      \\
        \bottomrule
    \end{tabular}
\end{table*}

\begin{table*}[h!]
\centering
\caption{gym-carla and evaluation Environment Parameters}
\begin{tabular}{ll}
\toprule
\textbf{Parameter} & \textbf{Value} \\
\midrule
Number of Vehicles & 200 \\
Number of Walkers & 0 \\
Random Seed & 1\\
Other Sensors Input\_names & lidar\_noground\\
Display Size & 400 \\
Max Past Step & 1 \\
Time Step (dt) & 0.1 \\
Discrete Control & False \\
Continuous Acceleration Range & [-3.0, 3.0] \\
Continuous Steering Range & [-0.2, 0.2] \\
Ego Vehicle Filter & vehicle.lincoln* \\
Traffic Manager Port & Random integer (2000 to 9000) \\
Town Map & town03 or town04 \\
Task Mode & Random \\
Max Time per Episode & 2000 \\
Max Waypoints & 12 \\
Observation Range & 32 \\
LiDAR Bin Size & 0.25 \\
Distance Behind Ego Vehicle & 12 \\
Lane Threshold & 2.0 \\
Desired Speed & 8 \\
Max Ego Vehicle Spawn Times & 200 \\
Display Route & True \\
PIXOR Grid Size & 64 \\
PIXOR Mode & False \\
Predict Speed & True \\
\bottomrule
\end{tabular}
\label{table:gym_carla_params}
\end{table*}

\begin{figure*}
    \centering
    \includegraphics[width=1\linewidth]{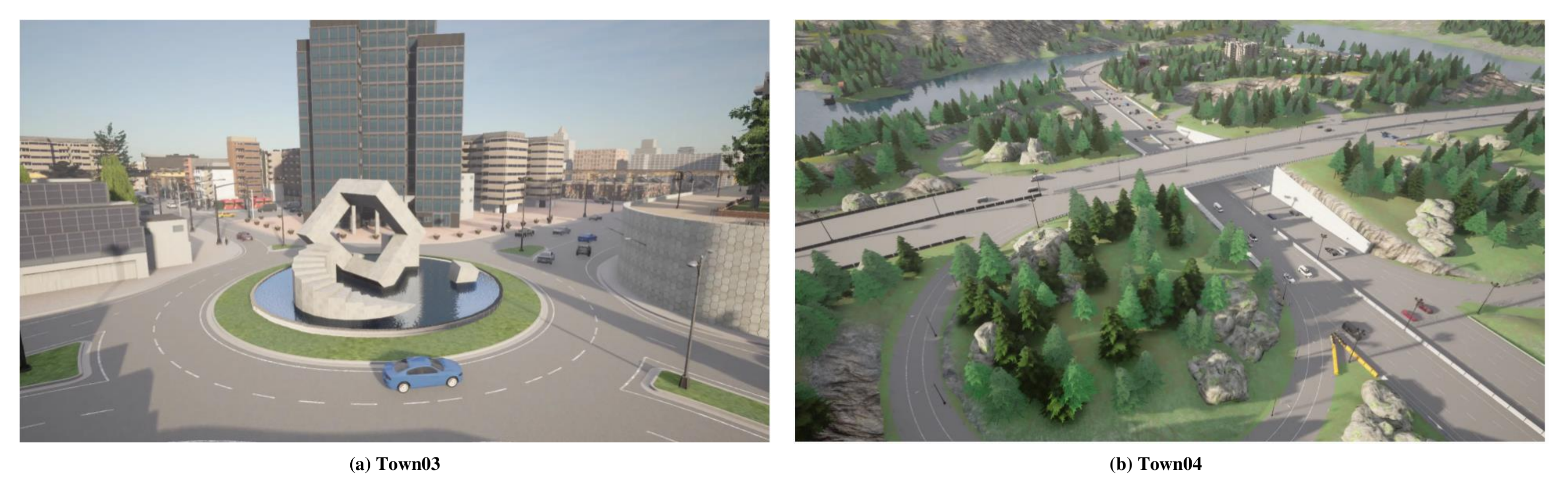}
    \caption{CARLA maps}
    \label{fig:enter-label}
\end{figure*}
\subsection{CARLA maps}
\label{sec:CARLA_maps}
To comprehensively evaluate the performance of MIMO-VLA, we utilized five maps in CARLA, including town03 and town04, as shown in Figure 6. Town03 is a complex urban map closely resembling real-world road environments, featuring tunnels, intersections, roundabouts, curves, and multi-turns. It covers an area of 400m $\times$ 400m, with a total road length of approximately 6km. In contrast, Town04 is a smaller town set against a backdrop of snow-capped mountains and conifers, with a multi-lane road forming a ``figure of 8" around the town.
\subsection{Training datasets}
\label{sec:traning_datasets}
We trained all comparison methods using an expert dataset $D_{expert}$, which is 5.69GB in size and contains 13,761 frames. Of this, 90\% was used for training and the remaining 10\% for testing. The comparison methods were evaluated online in CARLA town03 under random mode. Following DT \cite{chen2021decision}, we assessed sequence fusion performance for both single and multiple time steps. We set the context length $H = 1$, resulting in a fusion sequence length of 489 tokens, which includes 64 tokens for the 128$\times$128 image and padded text sequences of length 424 (including an empty placeholder token).

Due to computational constraints, we explored trajectory sequences with a maximum length of $489 \times 4=1956$ to assess performance over longer contexts. Further analysis of MIMO-VLA’s decision-making ability with longer trajectory contexts is provided in Appendix~\ref{apeedinx: long text}. We also evaluated performance across different modalities and generalization capabilities in town04. Detailed information on the CARLA maps can be found in Appendix~\ref{sec:CARLA_maps}.

All comparison methods were tested online in the CARLA simulator over 20 episodes, each consisting of 1,000 steps and 200 vehicles, with driving routes and traffic scenarios generated in random mode.
\subsection{Effect of Longer Trajectory Context on Decision-Making Ability}
\label{apeedinx: long text}
As shown in Table 9, increasing the context length $H$ of input trajectories provides some improvement in MIMO-VLA’s overall DS and AR, but the gains are modest. The improvement mainly stems from higher route completion rates and lower collision rates associated with longer time steps. For example, when $H = 4$, the sequence length extends to 1,956 tokens—a fourfold increase—yet the improvements in DS and AR remain limited. In fact, for metrics such as AR and ASD, the performance for $H = 4$ can even be worse than that for $H = 1$, suggesting that excessively long trajectories may introduce redundant information that can negatively impact decision-making.

This result highlights several key points. While longer contexts provide the model with more historical information, too much data can hinder its ability to filter and extract useful decision signals, leading to redundancy. Redundant information not only increases computational complexity but can also distract the model, reducing its capacity to capture critical features and negatively affecting overall decision-making. Consequently, shorter context sequences offer more concise and precise inputs, enabling quicker and more accurate judgments. These findings suggest that the current sequence fusion method has limited benefits from longer contexts. Prior work \cite{chen2021decision, hung2019optimizing} indicates that longer contexts can improve decision control, motivating future research on compressing historical information and efficiently fusing it to enhance decision-making.

\begin{table*}[t!]
\centering
\caption{Evaluation of MIMO-VLA with Longer Contexts for Multimodal Input in Town03 (Random)}
\resizebox{135mm}{!}{
\begin{tabular}{ccccc|cccc}
\toprule
Input& $\mathcal{L}_{\text{image}}$ & $H$ & DS $\uparrow$ & AR ($f$) $\uparrow$ & ASD(m) $\uparrow$  & ER(\%) $\uparrow$ & OR(\%) $\downarrow$ & CR(\%) $\downarrow$   \\
\midrule
image& $\times$ &1&$29.55\pm6.17$ & $226.91\pm42.24 $& $54.24\pm 4.30$& $11.85\pm 0.68$& $20.22\pm5.57$ & $70.00\pm10.5$\\
image& $\times$ &4&$22.38\pm4.96$ & $155.79\pm31.87$ &$ 32.45\pm1.74 $& $14.41\pm 0.59 $&$15.93\pm2.65$& $40.00\pm 11.23$\\
\midrule
text& $\times$ &1& $37.44\pm10.11$& $248.89\pm 52.91 $&$ 47.37\pm 5.43 $& $15.63\pm 1.98$& $17.02\pm 2.71 $& $40.00\pm11.24$\\
text& $\times$ & 4& $44.16\pm7.39$ & $252.10\pm38.94$ & $46.96\pm3.23$ &$15.66\pm 1.06 $& $12.86\pm2.45$& $60.00\pm 11.23$ \\
\midrule
image, text & $\times$ & 1 &$  68.10\pm13.20 $  & $ 417.24\pm57.41 $ &$ 58.81\pm 6.55$&$ 13.71\pm 1.26$ & $11.39\pm 2.41$ &$ 40.00\pm 11.24$\\
image, text & $\times$ & 4 & $74.85\pm10.97$  & $331.78\pm49.88$ &$50.63\pm4.73$& $18.62\pm1.95$& $15.96\pm 2.45$ & $25.00\pm9.93$\\
\midrule
image, text & $\checkmark$ & 1 &$92.78\pm23.75$ &$466.80\pm91.66$ &$71.77\pm9.40$ & $16.35\pm 1.56 $&$15.33\pm4.36 $ &$55.00\pm11.41$ \\
image, text & $\checkmark$ & 4&$105.25\pm14.03$  &$ 349.52\pm49.75$ &$59.76\pm\ 5.04$  &$25.02\pm2.57$ &$19.93\pm2.11$ &$30.00\pm10.51$\\

\bottomrule
\end{tabular}}
\label{tab:table5}
\end{table*}

\subsection{Reward function}
\label{sec:app-reward}
We use the default reward function of the Gym-Carla benchmark \citep{chen2019model} to evaluate all experimental methods, as follows:
\begin{equation}
\begin{aligned}
\label{eq:2-6}
\textstyle
f =200r_c+ v_{lon} + 10r_f + r_o\\ - 5 \alpha^2 + 0.2r_{lat} - 0.1\\
\end{aligned}
\end{equation}
where $r_c$ is the collision-related reward, set to $-1$ if the ego vehicle collides and $0$ otherwise. $v_{lon}$ denotes the longitudinal speed of the ego vehicle. $r_f$ is the reward for exceeding the desired speed ($8m/s$ in this case), set to $-1$ if the speed limit is exceeded and $0$ otherwise. $r_o$ is set to $-1$ if the ego vehicle leaves its lane, and $0$ otherwise. $\alpha$ represents the steering angle of the ego vehicle in radians. The lateral acceleration reward $r_{lat}$ is computed as $r_{lat} = -|\alpha| \cdot v_{lon}^2$. Finally, a constant term is added to prevent the ego vehicle from remaining stationary.
\subsection{Performance metrics of autonomous driving models}
\label{sec:metrics}
We use several key metrics to evaluate the performance of autonomous driving models across various driving scenarios:
\begin{itemize}
  \item Collision Rate (CR): The frequency at which the vehicle collides with obstacles or other vehicles, critical for assessing driving safety.
  \item Outlane Rate (OR): The rate at which the vehicle deviates from its designated lane, reflecting the model’s lane-keeping ability.
  \item Episode Completion Rate (ER): The percentage of driving tasks or episodes successfully completed, with higher values indicating better task performance.
  \item Average Safe Driving Distance (ASD): The average distance driven without incidents, such as collisions or off-road events, highlighting safe driving capability over extended periods.
  \item Average Return (AR): The cumulative reward collected during driving tasks, reflecting both task performance and adherence to safety guidelines.
  \item Driving Score (DS): A comprehensive metric capturing overall performance in terms of safety, efficiency, and traffic compliance, following the CARLA Leaderboard methodology.
\end{itemize}

For AR, we adopt the reward function $f$ from \citet{chen2019model}, which evaluates driving dynamics including yaw, collisions, speeding, and lateral velocity. Model selection prioritizes checkpoints optimizing both DS and AR. The remaining metrics (ER, OR, ASD) are implemented following \citet{gao2024enhance}.
\begin{equation}
 \textstyle
      CR = \frac{N_{\text{collisions}}}{N_{\text{total\_episodes}}}, OR = \frac{N_{\text{off\_road\_events}}}{N_{\text{total\_episodes}}}
\end{equation}
\begin{equation}
\textstyle
 ER = \frac{N_{\text{completed\_steps}}}{N_{\text{total\_steps}}},ASD = \frac{\sum_{i=1}^{N_{\text{episodes}}} \text{distance}_i}{N_{\text{total\_episodes}}}
\end{equation}

\begin{equation}
\textstyle
 AR = \frac{\sum_{i=1}^{N_{\text{episodes}}} \text{rewards}_i}{N_{\text{total\_episodes}}}, DS = ER \times AR
\end{equation}
where \(N_{\text{collisions}}\) denotes the number of collisions during the episode, and \(N_{\text{total\_episodes}}\) is the total number of episodes in the test. \(N_{\text{off\_road\_events}}\) represents the number of times the vehicle went off-road, while \(N_{\text{total\_steps}}\) is the total number of steps across all episodes. \(\text{distance}_i\) is the distance driven during the \(i\)-th safe driving episode, and \(N_{\text{safe\_episodes}}\) is the number of episodes completed without incidents, such as collisions or off-road events. \(N_{\text{completed\_steps}}\) is the number of successfully completed steps, and \(N_{\text{total\_steps}}\) is the total number of steps in the episode. Finally,\(AR\) denotes the average reward $f$ collected during the episode.
\subsection{The natural language template for text input}
\label{sec:template}
We collected information from the CARLA environment using additional sensors (e.g., speed and position sensors), excluding the ego vehicle’s acceleration and steering actions. This information is then converted into a natural language template interpretable by the VLA, as illustrated below:\\
$<$\textit{lateral\_dis, delta\_yaw, speed, vehicles\_info}$>$ = $<$\textit{observation\_vehicle\_state}$>$\\
$<$\textit{vehicles\_num}$>$ = $<$\textit{len(vehicles\_info)}$>$\\
$<$\textit{multi\_dis += str(vehicles\_info[i][0])+"", multi\_yaw += str(vehicles\_info[i][1])+"", multi\_speed += str(vehicles\_info[i][2])+""}$>$\\
$<$\textit{\textbf{if vehicles\_num}=\textit{1:}}$>$\\
$<$\textit{\textbf{new\_input}="You can see that there is a car. Its speed, straight-line distance from you, and angle in the direction you're heading are respectively \{\textbf{multi\_speed}\} m/s, \{\textbf{multi\_dis}\} m, \{\textbf{multi\_yaw}\}°." "You are now \{\textbf{lateral\_dis}\}m laterally away from your driving route. "}$>$\\
$<$\textit{\textbf{elif vehicles\_num}$>$\textit{1:}}$>$\\
$<$\textit{\textbf{new\_input}="You can see that there are {\textbf{vehicles\_num}} cars. Their speed, straight-line distance from you, and angle in the direction you're heading are respectively \{\textbf{multi\_speed}\} m/s, \{\textbf{multi\_dis}\} m, \{\textbf{multi\_yaw}\}°." "You are now \{\textbf{lateral\_dis}\}m laterally away from your driving route. "}$>$\\
$<$\textit{\textbf{elif vehicles\_num}}=\textit{0:}$>$\\
$<$\textit{\textbf{new\_input}="You see no car here, and you are now \{\textbf{lateral\_dis}\}m laterally away from your driving route."}$>$\\

We followed \citet{wang2023chatgpt} \textit{"Is ChatGPT a Good NLG Evaluator?}" approach. The complete evaluation prompt template for using GPT-4o \cite{openai2023gpt4} is as follows:"\textit{The document contains 50 similar examples as described above. For each example, based on the given Input\_0: and Question\_0:, please evaluate and score the responses generated by the five methods (MIMO-VLA\_pred\_0, Openvla\_pred\_0, DriverGPT4\_pred\_0, MIMO-VLA\_image\_pred\_0, and MIMO-VLA\_language\_pred\_0) using a 10-point scale with the following criteria:Not Acceptable ($<3$), Acceptable ($3\leq \text{score}< 6$), and Good ($\geq 6$). Please output the individual scores for each example. After evaluating all 50 examples, calculate the average rates for: Not Acceptable, Acceptable, Good, and Excellent performance for each method.}"\\

\subsection{The benefits of cross-entropy loss and label smoothing loss for MIMO-VLA}
\label{sec:overfitting}

\begin{figure*}
    \centering
    \includegraphics[width=1\linewidth]{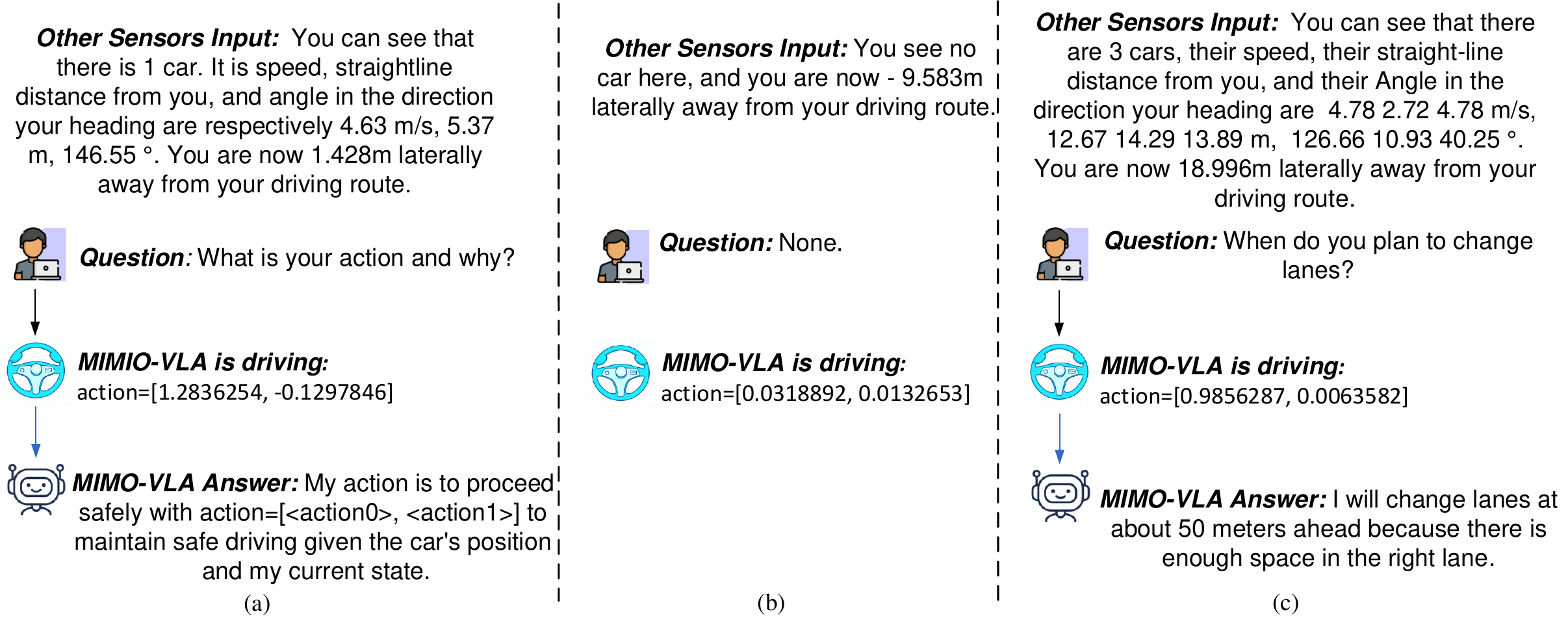}
     \caption{An example illustrating how MIMO-VLA seamlessly engages in conversation with a human while simultaneously making real-time driving decisions.}
    \label{fig:enter-label}
\end{figure*}

We observed that simply replacing specific numerical values in the translation template \cite{chen2024driving} produces minimal representational differences due to the sequential nature of the data. This makes conventional cross-entropy loss prone to overfitting in text generation tasks. As shown in Table 10, evaluations on both town03 and town04 revealed a decline in the model’s decision-making performance when using standard cross-entropy loss. In contrast, cross-entropy loss with smoothed labels yielded better performance. Consequently, we adopted cross-entropy loss with smoothed labels for text generation in MIMO-VLA in our experiments.
\subsection{Impact of Training Data on Model Decision Performance}
In the multimodal ablation experiments on MIMO-VLA, summarized in Table 11, we systematically removed or replaced individual modalities to evaluate their contributions to decision-making. The results show that models using both image and text inputs significantly outperform those using only a single modality in terms of decision accuracy and stability. This indicates that the text modality provides higher-level semantic information that complements visual inputs, thereby enhancing overall decision-making.

Furthermore, as shown in Table 11, models with text-only input outperform those with image-only input. This highlights that the text modality in our dataset—particularly the “other sensors input” shown in Figure 6—offers highly informative cues that substantially improve the model’s decision-making capability.

\begin{table*}[t!]
\centering
\caption{Evaluation of MIMO-VLA Performance Using Smoothed Label Loss and Cross-Entropy Loss (\textit{H=4})}
\resizebox{135mm}{!}{
\begin{tabular}{cccc|cccc}
\toprule
 $\mathcal{L}_{\text{language}}$& Town& DS $\uparrow$ & AR ($f$) $\uparrow$  & ASD(m) $\uparrow$ & ER(\%) $\uparrow$ & OR(\%) $\downarrow$ & CR(\%) $\downarrow$  \\
\midrule
Cross Entropy& town03 & $48.97\pm 7.60$ &$ 296.53\pm40.72$ & $47.10\pm4.87\ $& $15.37\pm 0.85$ & $12.41\pm2.73$&$ 35.00\pm10.94$ \\
Smooth Label& town03 &$105.25\pm14.03$  & $349.52\pm49.75 $& $59.76\pm\ 5.04$  &$25.02\pm2.57$ &$19.93\pm2.11$ &$30.00\pm10.51$\\
\midrule
Cross Entropy& town04 &$ 66.69\pm 16.97$ & $358.11\pm61.10$ & $52.72\pm5.44$&$ 15.43\pm 1.11$ & $9.63\pm1.42$& $55.00\pm 11.41$ \\
Smooth Label& town04 & $94.26\pm15.26$ &$ 384.52\pm51.72 $&$56.93\pm4.03 $ &$21.49\pm1.86$ &$12.75\pm2.28$ &$45.00\pm11.41$\\
\bottomrule
\end{tabular}}
\label{tab:example}
\end{table*}

\begin{table*}[t!]
\centering
\caption{Impact of Different Modal Inputs on MIMO-VLA Decision-Making in Town03 (Random) with \textit{H=4}}
\resizebox{135mm}{!}{
\begin{tabular}{ccccc|cccc}
\toprule
Input& $\mathcal{L}_{\text{image}}$  & DS $\uparrow$ & AR ($f$) $\uparrow$ & ASD(m) $\uparrow$  & ER(\%) $\uparrow$ & OR(\%) $\downarrow$ & CR(\%) $\downarrow$   \\
\midrule
image& $\times$ &$22.38\pm4.96$ & $155.79\pm31.87$ & $32.45\pm1.74$ & $14.41\pm 0.59$ &$15.93\pm2.65$& $40.00\pm 11.23$\\
text& $\times$ & $44.16\pm7.39$& $252.10\pm38.94$ &$ 46.96\pm3.23 $&$15.66\pm 1.06$ & $12.86\pm2.45$& $60.00\pm 11.23$ \\
image, text & $\times$  & $74.85\pm10.97$  & $331.78\pm49.88 $&$50.63\pm4.73$&$ 18.62\pm1.95$& $15.96\pm 2.45$ & $25.00\pm9.93$\\
image, text & $\checkmark$ &$105.25\pm14.03  $& $349.52\pm49.75$ &$59.76\pm\ 5.04$  &$25.02\pm2.57$ &$19.93\pm2.11$ &$30.00\pm10.51$\\

\bottomrule
\end{tabular}}
\label{tab:table5}
\end{table*}
\subsection{The noise datasets}
\label{sec:noise input}
The noise consisted of information entirely unrelated to the current driving scenario, as follows:
\{"A playful puppy brings joy and laughter to our days",
"The whisper of the wind carries secrets of the universe",
"A hidden garden blooms with the magic of nature's colors",
"The aroma of fresh coffee awakens the senses each morning",
"A handwritten letter feels like a warm hug from afar",
"The glimmer of fireflies creates a magical summer night",
"A spontaneous adventure can lead to unforgettable memories",
"The serenity of a quiet lake reflects the beauty of the world",
"A gentle touch can convey love without a single word",
"The laughter of friends is the sweetest melody of all",
"A warm hug is a universal language of comfort",
"The dance of leaves in the breeze tells stories of change",
"A cozy fire invites stories and shared moments",
"The beauty of art inspires creativity and self-expression",
"A day spent volunteering fills the heart with purpose",
"The excitement of a new book is like embarking on a journey",
"A delicious meal shared brings people closer together",
"The sound of laughter can brighten even the gloomiest day",
"A fleeting moment can hold the weight of a thousand memories",
"The charm of small towns lies in their simple beauty",
"A gentle rain nurtures the earth and inspires growth",
"A colorful painting captures the essence of joy",
"The peace of a mountain retreat refreshes the soul",
"A favorite mug holds warmth and comfort on a chilly day",
"The rustle of leaves underfoot reminds us of nature's rhythm",
"A well-crafted story has the power to transport us anywhere",
"The thrill of discovery keeps our spirits young and curious",
"A cherished photograph holds a lifetime of memories",
"The beauty of winter blankets the world in quiet calm",
"A moment of kindness can change the trajectory of a day",
"The aroma of spices fills the kitchen with warmth and love",
"A shared joke creates bonds that laughter alone cannot",
"The glow of a sunrise fills the heart with hope",
"A melody can linger in the mind long after it fades",
"The colors of autumn leaves create a vibrant tapestry",
"A soft pillow cradles the head and invites sweet dreams",
"The laughter of children brings joy and light to our lives",
"A surprise visit from a friend can brighten any day",
"The beauty of a flower garden is a celebration of life",
"A good book can be a loyal companion on lonely nights",
"The embrace of nature can heal and rejuvenate the spirit",
"A treasure hunt ignites the spirit of adventure",
"The warmth of homemade cookies fills the home with love",
"A playful kitten brings joy and mischief to our lives",
"The scent of pine trees evokes memories of the forest",
]\}.
\begin{figure*}
    \centering
    \includegraphics[width=1\linewidth]{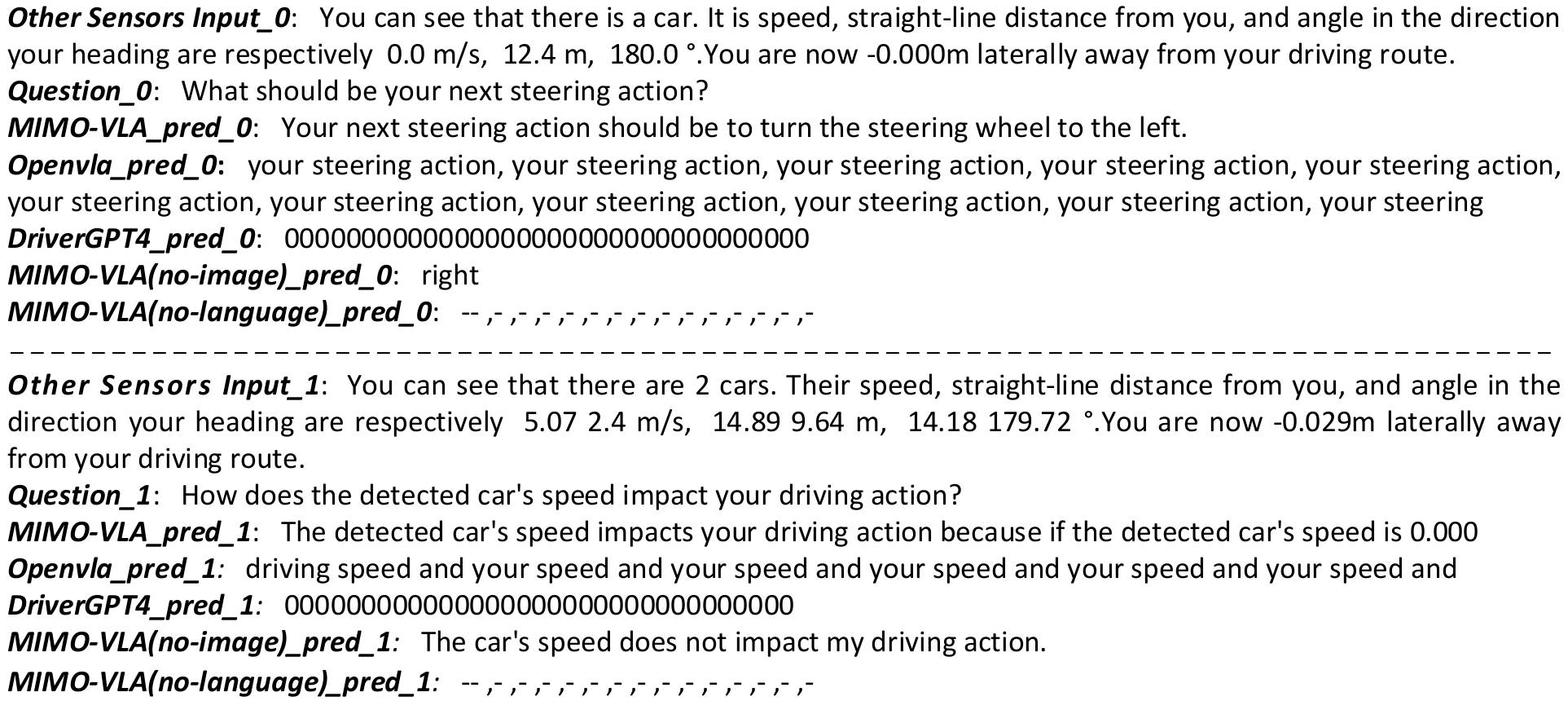}
     \caption{An example illustrating how MIMO-VLA seamlessly engages in conversation with a human while simultaneously making real-time driving decisions.}
    \label{fig:enter-label}
\end{figure*}
Additionally, we generated a dataset of over 500 question-answer pairs using GPT to introduce noise into the QA content. A random subset is shown below:
\{
        "question": "What is the national flower of Egypt?",
        "answer": "The national flower of Egypt is the lotus."
    \},
    \{
        "question": "How do I make a vegetable soup?",
        "answer": "To make a vegetable soup, chop vegetables, simmer them in broth, and season with herbs and spices."
    \},
    \{
        "question": "What is the largest gulf in the world?",
        "answer": "The largest gulf in the world is the Gulf of Mexico."
    \},
    \{
        "question": "How do I clean my humidifier?",
        "answer": "To clean your humidifier, empty it, wash the tank with soap and water, and rinse thoroughly."
    \},
    \{
        "question": "What is the national animal of Egypt?",
        "answer": "The national animal of Egypt is the steppe eagle."
    \},
    \{
        "question": "How do I make a cup of lavender tea?",
        "answer": "To make a cup of lavender tea, steep dried lavender flowers in hot water for 5-7 minutes."
    \},
    \{
        "question": "What is the largest palace in the world?",
        "answer": "The largest palace in the world is the Palace of the Parliament in Romania."
    \},
    \{
        "question": "How do I check my computer's hard drive health?",
        "answer": "To check your computer's hard drive health, use a diagnostic tool like CrystalDiskInfo or HD Tune."
    \},
    \{
        "question": "What is the national flower of Greece?",
        "answer": "The national flower of Greece is the violet."
    \},
    \{
        "question": "How do I make a chicken soup?",
        "answer": "To make a chicken soup, simmer chicken, vegetables, and broth together, and season with herbs and spices."
    \},
    \{
        "question": "What is the largest strait in the world?",
        "answer": "The largest strait in the world is the Strait of Malacca."
    \},
    \{
        "question": "How do I clean my dehumidifier?",
        "answer": "To clean your dehumidifier, empty it, wash the tank with soap and water, and rinse thoroughly."
    \},
    \{
        "question": "What is the national animal of Greece?",
        "answer": "The national animal of Greece is the dolphin."
    \},
    \{
        "question": "How do I make a cup of rose tea?",
        "answer": "To make a cup of rose tea, steep dried rose petals in hot water for 5-7 minutes."
    \},
    \{
        "question": "What is the largest castle in the world?",
        "answer": "The largest castle in the world is Prague Castle in the Czech Republic."
    \},
    \{
        "question": "How do I check my computer's network speed?",
        "answer": "To check your computer's network speed, use an online speed test tool and follow the instructions."
    \},
    \{
        "question": "What is the national flower of Turkey?",
        "answer": "The national flower of Turkey is the tulip."
    \},
    \{
        "question": "How do I make a beef stew?",
        "answer": "To make a beef stew, simmer beef, vegetables, and broth together, and season with herbs and spices."
    \},
    \{
        "question": "What is the largest canal in the world?",
        "answer": "The largest canal in the world is the Grand Canal in China."
    \},
    \{
        "question": "How do I clean my air purifier?",
        "answer": "To clean your air purifier, remove and wash the filters, and wipe the exterior with a damp cloth."
    \},
    \{
        "question": "What is the national animal of Turkey?",
        "answer": "The national animal of Turkey is the gray wolf."
    \},

    \{
        "question": "How do I make a cup of turmeric tea?",
        "answer": "To make a cup of turmeric tea, steep fresh turmeric slices in hot water for 5-7 minutes."
    \}
\subsection{Definition of the 50-Question Set}
\label{sec: problems set}
Questions are randomly selected from the following set of 50 questions:  \{"What are you seeing/observing?",
    "What are you paying attention to and why?",
    "Are there any traffic lights? What’s the color of the traffic light?",
    "What’s your current speed and steering angle?",
    "What is your action and why?",
    "Summarize the current driving scenario at a high level.",
    "How are you going to drive in this situation and why?",
    "What’s the straight-line distance to the nearest car?",
    "What is the angle of the nearest car relative to your heading?",
    "Is there any lateral deviation from your driving route?",
    "What should be your next steering action?",
    "What should be your next acceleration command?",
    "Is there any moving object around you?",
    "Describe the position of the car relative to your heading.",
    "What is your current lateral position relative to your route?",
    "What would be a safe driving action given the detected car's details?",
    "What is the speed of the detected car?",
    "How far is the detected car from you?",
    "What angle should you adjust your steering to avoid collision?",
    "Why is it important to note the angle of the detected car?",
    "Is the detected car in motion?",
    "What should you be cautious of given the car’s position?",
    "What action should be taken to maintain alignment on your driving route?",
     What should you avoid in this situation to prevent collision?,
    "What considerations are necessary for the detected car's speed?",
    "What’s the importance of your current lateral position in planning the next action?",
    Why did you brake just now?,
    "What factors are influencing your next driving decision?",
    "Is there any obstacle directly ahead?",
    "How should you interpret the car’s angle for your steering decision?",
    "What immediate adjustments are necessary for safe driving?",
    "How does the detected car's speed impact your driving action?",
    "What should be your focus given the detected car’s proximity and angle?",
    "What safe action is suggested based on the current scenario?",
    "What should you avoid in this situation to prevent collision?",
    "Is there a need for a speed adjustment?",
    "How will your steering angle change based on the detected car’s angle?",
    "What should you consider for maintaining a safe path?",
    "How would you describe the current traffic conditions?",
    "What immediate action is necessary given your current lateral position?",
    "What factors need to be monitored to ensure safe navigation?",
    "Is the detected car influencing your path directly?",
    "What is the priority in adjusting your speed and direction?",
    "\quad"\}

\subsection{Loss weights}
\label{sec: loss weights}
Our loss weighting analysis yielded several key insights. First, increasing the image reconstruction weight (from 0.5 to 2) degraded decision-making performance while leaving language generation largely unaffected. Second, increasing the language loss weight (from 0.1 to 10) significantly enhanced conversational quality but substantially impaired driving performance. Third, raising the action prediction weight (from 0.1 to 10) markedly improved driving performance with minimal impact on language outputs. Based on these observations, we identified the optimal weighting scheme for autonomous driving applications as: language loss weight = 0.1, action loss weight = 10, and image reconstruction weight = 0.5. This configuration provides the best balance between driving performance and auxiliary task capabilities.
\subsection{Further results on MIMO-VLA in answering questions}
\textbf{\textit{Other Sensors Input\_33}}: You can see that there is a car. Its speed, straight-line distance from you, and angle in the direction you're heading are respectively  2.45 m/s,  9.73 m,  179.82 °.You are now -0.009m laterally away from your driving route. \\
\textbf{\textit{Question\_33}}: What factors need to be monitored to ensure safe navigation?\\
\textbf{\textit{MIMO-VLA\_pred\_33}}: You need to monitor the speed, straight-line distance from you, and angle in the direction you're heading are respectively  2.45 m/s,  9.73 m,  179.82 °.\\
\textbf{\textit{Openvla\_pred\_33}}: monitored to safe to ensure safe and to ensure safe
monitored to safe\\
\textbf{\textit{DriverGPT4\_pred\_33}}: 00000000000000000000000000000000000000
\\
\textbf{\textit{MIMO-VLA\_image\_pred\_33}}: speed, straight-line distance from you, and angle in the direction you're heading are respectively  2.45 m/s,  9.73 m,  179.82 °.You are now -0.009m laterally away from your driving route.\\
\textbf{\textit{MIMO-VLA\_language\_pred\_33}}:
-
\rule{0.48\textwidth}{0.5pt}\\
\textbf{\textit{Other Sensors Input\_34}}: You can see that there is a car. Its speed, straight-line distance from you, and angle in the direction you're heading are respectively  2.49 m/s,  9.69 m,  179.81 °.You are now -0.011m laterally away from your driving route. \\
\textbf{\textit{Question\_34}}: What should be your next steering action?\\
\textbf{\textit{MIMO-VLA\_pred\_34}}: Your next steering action should be to turn the steering wheel to the left.\\
\textbf{\textit{Openvla\_pred\_34}}: your steering action and your next steering action and your steering action and your steering action and your steering action and \\
\textbf{\textit{DriverGPT4\_pred\_34}}: 0000000000000000000000000000000000000\\
\textbf{\textit{MIMO-VLA\_image\_pred\_34}}: right\\
\textbf{\textit{MIMO-VLA\_language\_pred\_34}}:
-\\
\rule{0.48\textwidth}{0.5pt}\\
\textbf{\textit{Other Sensors Input\_35}}: You can see that there is a car. Its speed, straight-line distance from you, and angle in the direction you're heading are respectively  2.47 m/s,  9.66 m,  179.79 °.You are now -0.012m laterally away from your driving route. \\
\textbf{\textit{Question\_35}}: What’s the importance of your current lateral position in planning the next action?\\
\textbf{\textit{MIMO-VLA\_pred\_35}}: The current lateral position is important in planning the next action because if you are too far away from your driving route.\\
\textbf{\textit{Openvla\_pred\_35}}: your current position in planning the next action \\
\textbf{\textit{DriverGPT4\_pred\_35}}: 000000000000000000000000000000000000\\
\textbf{\textit{MIMO-VLA\_image\_pred\_35}}: Your current lateral position is important in planning the next action because if you don’t plan the next action, you will not be able to reach your destination.\\
\textbf{\textit{MIMO-VLA\_language\_pred\_35}}:
-\\
\rule{0.48\textwidth}{0.5pt}\\
\textbf{\textit{Other Sensors Input\_36}}: You can see that there is a car. Its speed, straight-line distance from you, and angle in the direction you're heading are respectively  2.41 m/s,  9.64 m,  179.78 °.You are now -0.014m laterally away from your driving route. \\
\textbf{\textit{Question\_36}}: What safe action is suggested based on the current scenario?\\
\textbf{\textit{MIMO-VLA\_pred\_36}}: You can see that there is a car. Its speed, straight-line distance from you, and angle in the direction you're heading are respectively.\\
\textbf{\textit{Openvla\_pred\_36}}: suggested based on the current safe action on the current\\
\textbf{\textit{DriverGPT4\_pred\_36}}: 0000000000000000000000000000000000000\\
\textbf{\textit{MIMO-VLA\_image\_pred\_36}}: The car is speed, straight-line distance from you, and angle in the direction you're heading are respectively  2.41 m/s,  9.64 m,  179.78 °.You are now -0.014m laterally away from your driving route.\\
\textbf{\textit{MIMO-VLA\_language\_pred\_36}}:
-
-\\
\rule{0.48\textwidth}{0.5pt}\\
\textbf{\textit{Other Sensors Input\_37}}: You can see that there is a car. Its speed, straight-line distance from you, and angle in the direction you're heading are respectively  2.37 m/s,  9.63 m,  179.77 °.You are now -0.016m laterally away from your driving route. \\
\textbf{\textit{Question\_37}}: Is there any lateral deviation from your driving route?\\
\textbf{\textit{MIMO-VLA\_pred\_37}}: There is no lateral deviation from your driving route. \\
\textbf{\textit{Openvla\_pred\_37}}: driving route from your route and your route from your
driving route from your route from your\\
\textbf{\textit{DriverGPT4\_pred\_37}}: 0000000000000000000000000000000000000\\
\textbf{\textit{MIMO-VLA\_image\_pred\_37}}: Yes, there is lateral deviation from your driving route.\\
\textbf{\textit{MIMO-VLA\_language\_pred\_37}}:
-\\
\rule{0.48\textwidth}{0.5pt}\\
\textbf{\textit{Other Sensors Input\_38}}: You can see that there is a car. Its speed, straight-line distance from you, and angle in the direction you're heading are respectively  2.37 m/s,  9.63 m,  179.76 °.You are now -0.018m laterally away from your driving route. \\
\textbf{\textit{Question\_38}}: How will your steering angle change based on the detected car’s angle?\\
\textbf{\textit{MIMO-VLA\_pred\_38}}: Your steering angle will change based on the detected car’s angle.\\
\textbf{\textit{Openvla\_pred\_38}}: your steering on the detected car’s steering \\
\textbf{\textit{DriverGPT4\_pred\_38}}: 0000000000000000000000000000000000\\
\textbf{\textit{MIMO-VLA\_image\_pred\_38}}: The steering angle will change based on the detected car’s angle.\\
\rule{0.48\textwidth}{0.5pt}\\
\textbf{\textit{Other Sensors Input\_39}}: You can see that there is a car. Its speed, straight-line distance from you, and angle in the direction you're heading are respectively  2.41 m/s,  9.63 m,  179.75 °.You are now -0.020m laterally away from your driving route. \\
\textbf{\textit{Question\_39}}: Is there any moving object around you?\\
\textbf{\textit{MIMO-VLA\_pred\_39}}: Yes, there is a car. Its speed, straight-line distance from you, and angle in the direction you're heading are respectively  2.\\
\textbf{\textit{Openvla\_pred\_39}}: moving object of you and your work with your work?\\
\textbf{\textit{DriverGPT4\_pred\_39}}: 0000000000000000000000000000000000000\\
\textbf{\textit{MIMO-VLA\_image\_pred\_39}}: Yes, there is a car.\\
\textbf{\textit{MIMO-VLA\_language\_pred\_39}}:
-
-\\
\rule{0.48\textwidth}{0.5pt}\\
\textbf{\textit{Other Sensors Input\_40}}: You can see that there is a car. Its speed, straight-line distance from you, and angle in the direction you're heading are respectively  2.45 m/s,  9.63 m,  179.74 °.You are now -0.022m laterally away from your driving route. \\
\textbf{\textit{Question\_40}}: Is there any moving object around you?\\
\textbf{\textit{MIMO-VLA\_pred\_40}}: Yes, there is a car. Its speed, straight-line distance from you, and angle in the direction you're heading are respectively  2.\\
\textbf{\textit{Openvla\_pred\_40}}: moving object of you and your work with your work?
\\
\textbf{\textit{DriverGPT4\_pred\_40}}: 00000000000000000000000000000000\\
\textbf{\textit{MIMO-VLA\_image\_pred\_40}}: Yes, there is a car.
\\
\textbf{\textit{MIMO-VLA\_language\_pred\_40}}:
-
-\\
\rule{0.48\textwidth}{0.5pt}\\
\textbf{\textit{Other Sensors Input\_41}}: You can see that there is a car. Its speed, straight-line distance from you, and angle in the direction you're heading are respectively  2.46 m/s,  9.63 m,  179.73 °.You are now -0.024m laterally away from your driving route. \\
\textbf{\textit{Question\_41}}: How does the detected car's speed impact your driving action?\\
\textbf{\textit{MIMO-VLA\_pred\_41}}: The detected car's speed impacts your driving action because if the detected car's speed is faster than your driving speed\\
\textbf{\textit{Openvla\_pred\_41}}: speed with your speed and your speed with your speed
speed with your speed with your speed with your speed with your speed with\\
\textbf{\textit{DriverGPT4\_pred\_41}}: 0000000000000000000000000000000000000\\
\textbf{\textit{MIMO-VLA\_image\_pred\_41}}: The car's speed impacts my driving action because the car's speed is increasing, which means that the car's speed is getting faster\\
\textbf{\textit{MIMO-VLA\_language\_pred\_41}}:
-
-\\
\rule{0.48\textwidth}{0.5pt}\\
\textbf{\textit{Other Sensors Input\_42}}: You can see that there is a car. Its speed, straight-line distance from you, and angle in the direction you're heading are respectively  2.43 m/s,  9.63 m,  179.73 °.You are now -0.027m laterally away from your driving route. \\
\textbf{\textit{Question\_42}}: What is your action and why?\\
\textbf{\textit{MIMO-VLA\_pred\_42}}: My action is to turn left. I am turning left because I want to turn left.\\
\textbf{\textit{Openvla\_pred\_42}}: “ your action” and your action and your
“ your action” and your action and your\\
\textbf{\textit{DriverGPT4\_pred\_42}}: 0000000000000000000000000000000000000\\
\textbf{\textit{MIMO-VLA\_image\_pred\_42}}: My action is straight-line distance from you, and why is because I am going straight.\\
My action is straight-line distance from you, and why is because I am going straight.
My action is straight-line distance from you, and why is because I am going straight.\\
\textbf{\textit{MIMO-VLA\_language\_pred\_42}}:
-
-\\
\rule{0.48\textwidth}{0.5pt}\\
\textbf{\textit{Other Sensors Input\_43}}: You can see that there are 2 cars. Their speed, straight-line distance from you, and angle in the direction you're heading are respectively  5.07 2.4 m/s,  14.89 9.64 m,  14.18 179.72 °.You are now -0.029m laterally away from your driving route. \\
\textbf{\textit{Question\_43}}: What safe action is suggested based on the current scenario?\\
\textbf{\textit{MIMO-VLA\_pred\_43}}: What safe action is suggested based on the current scenario?
What safe action is suggested based on the current scenario?
What safe action is suggested based\\
\textbf{\textit{Openvla\_pred\_43}}: What safe action is based on the current?
What safe action is based on the current\\
\textbf{\textit{DriverGPT4\_pred\_43}}: 0000000000000000000000000000000000000\\
\textbf{\textit{MIMO-VLA\_image\_pred\_43}}: straight-line distance from you, and angle in the direction you're heading are respectively  5.07 2.4 m/s,  14.89 9.64 m,  14.18 179.72 °.\\
\textbf{\textit{MIMO-VLA\_language\_pred\_43}}:
-
-
- ,-
- ,-
\\
\rule{0.48\textwidth}{0.5pt}\\
\textbf{\textit{Other Sensors Input\_44}}: You can see that there are 2 cars. Their speed, straight-line distance from you, and angle in the direction you're heading are respectively  4.98 2.38 m/s,  14.15 9.66 m,  14.95 179.71 °.You are now -0.032m laterally away from your driving route. \\
\textbf{\textit{Question\_44}}: Is the detected car in motion?\\
\textbf{\textit{MIMO-VLA\_pred\_44}}: Yes
answer:
No
answer:
Yes
answer:
No
answer\\
\textbf{\textit{Openvla\_pred\_44}}: motion in the detected car in motion in the detected car in motion in the detected car in motion in the detected car in motion in the detected car in motion in the detected car in motion in the detected car in motion in the detected car in motion in the detected car in motion in the detected car in motion in the\\
\textbf{\textit{DriverGPT4\_pred\_44}}: 00000000000000000000000000000000000000\\
\textbf{\textit{MIMO-VLA\_image\_pred\_44}}: Yes, the detected car is in motion.
""""""""""""""""""""""\\
\textbf{\textit{MIMO-VLA\_language\_pred\_44}}:
-
-\\
\rule{0.48\textwidth}{0.5pt}\\
\end{document}